\documentclass[a4paper]{article}

\usepackage{authblk}
\usepackage[utf8]{inputenc}
\usepackage[T1]{fontenc}
\usepackage{pslatex}
\usepackage[english]{babel}
\usepackage{fancyhdr}
\usepackage{hyperref}
\usepackage{amsmath,amssymb,amsfonts}
\usepackage{graphicx}
\usepackage{amsmath}
\usepackage{amsthm}
\usepackage{amsfonts}
\usepackage{mathtools}
\usepackage{amssymb}
\usepackage{algorithm}
\usepackage[noEnd=false]{algpseudocodex}
\usepackage{microtype}
\usepackage{rotating}
\usepackage{todonotes}
\usepackage{tikz}
\usepackage{xparse}
\usepackage{appendix}
\usepackage{url}
\usepackage{tabularx}
\usepackage{paralist}
\usepackage{booktabs}
\usepackage{multirow}
\usepackage{subcaption}
\usepackage{cancel}
\usetikzlibrary{calc,shapes.callouts,shapes.arrows}
\usetikzlibrary{automata}
\usetikzlibrary{positioning}
\usetikzlibrary{decorations.pathreplacing}
\usetikzlibrary{decorations.pathmorphing}
\usetikzlibrary{arrows}
\usetikzlibrary{shapes}

\newcommand{\F}{\mathbb{F}}

\newtheorem*{problem*}{Problem}

\DeclarePairedDelimiter\floor{\lfloor}{\rfloor}

\providecommand{\keywords}[1]{\textbf{\textit{Keywords }} #1}

\begin{document}

\title{A New Angle: On Evolving Rotation Symmetric Boolean Functions}

\author[1,2]{Claude Carlet}
\author[3]{Marko \DJ urasevic}
\author[3]{Bruno Ga\v{s}perov}
\author[3]{Domagoj Jakobovic}
\author[4]{Luca Mariot}
\author[5]{Stjepan Picek}

\affil[1]{{\normalsize Department of Mathematics, Universit\'{e} Paris 8, 2 rue de la libert\'{e}, 93526 Saint-Denis Cedex, France}}

\affil[2]{{\normalsize University of Bergen, Bergen, Norway} \\
	
	{\small \texttt{claude.carlet@gmail.com}}}

\affil[3]{{\normalsize Faculty of Electrical Engineering and Computing, University of Zagreb, Unska 3, Zagreb, Croatia} \\

{\small \texttt{marko.durasevic@fer.hr, bruno.gasperov@fer.hr, domagoj.jakobovic@fer.hr}}}

\affil[4]{{\normalsize Semantics, Cybersecurity and Services Group, University of Twente, 7522 NB Enschede, The Netherlands} \\
	
	{\small \texttt{l.mariot@utwente.nl}}}

\affil[5]{{\normalsize Digital Security Group, Radboud University, Postbus 9010, 6500 GL Nijmegen, The Netherlands} \\
	
	{\small \texttt{stjepan.picek@ru.nl}}}
	
\maketitle

\begin{abstract}
Rotation symmetric Boolean functions represent an interesting class of Boolean functions as they are relatively rare compared to general Boolean functions. At the same time, the functions in this class can have excellent properties, making them interesting for various practical applications.
The usage of metaheuristics to construct rotation symmetric Boolean functions is a direction that has been explored for almost twenty years. Despite that, there are very few results considering evolutionary computation methods. 
This paper uses several evolutionary algorithms to evolve rotation symmetric Boolean functions with different properties. Despite using generic metaheuristics, we obtain results that are competitive with prior work relying on customized heuristics. Surprisingly, we find that bitstring and floating point encodings work better than the tree encoding. Moreover, evolving highly nonlinear general Boolean functions is easier than rotation symmetric ones. 
\end{abstract}


\section{Introduction}
\label{sec:intro}

Boolean functions are mathematical objects with various applications, including cryptography~\cite{Meaux2016}, combinatorics~\cite{Rothaus}, coding theory~\cite{1056589,KERDOCK1972182}, sequences~\cite{1056589}, telecommunications~\cite{Paterson}, and computational complexity theory~\cite{10.5555/1540612}.
Naturally, for Boolean functions to be useful across various applications, they must fulfill various properties, such as being balanced and exhibiting high nonlinearity.
Finding Boolean functions with specific properties can be rather difficult, which is why the research community has been actively investigating the design of Boolean functions for nearly 50 years.
In that respect, approaches to constructing Boolean functions can be divided into algebraic construction and various search techniques.\footnote{Some works also combine theory and search techniques, e.g.,~\cite{4167738,DBLP:conf/ppsn/PicekMBJ14}}. Within search techniques, the most common division is into random search and metaheuristics.
Unfortunately, sometimes even those approaches are not sufficient due to the vast number of Boolean functions of $n$ inputs, which is equal to $2^{2^n}$ (see Table~\ref{tab:nr_boolean}). Clearly, for $n=6$, an exhaustive search already becomes impossible. In such cases, it might be beneficial to focus on special classes of Boolean functions that are smaller and, thus, more amenable to search and enumeration but still large enough to contain many interesting functions. One such class is rotation symmetric Boolean functions - those functions that are invariant under cyclic shifts of the input coordinates. These functions have played a pivotal role in surpassing the quadratic bound.

The initial motivation for studying rotation-symmetric Boolean functions can be traced back to the reason above: this class is significantly smaller than the class of general Boolean functions while still containing a large number of interesting functions. Moreover, such functions have a nice structure and allow for a compact representation~\cite{Mesnager2016}. We provide comparisons of class sizes for general Boolean functions, bent functions, and rotation symmetric functions in Table~\ref{tab:nr_boolean}.
Finally, the class of rotation symmetric Boolean functions is rich with cryptographically significant Boolean functions. For instance, Kavut et al. found Boolean functions in 9 variables with nonlinearity 241~\cite{4167738}. This achievement resolved an almost three-decade-old open problem and, notably, was accomplished using heuristics.

\begin{table}
\setlength{\tabcolsep}{2.3pt}
\scriptsize
  \centering
  \caption{The number of Boolean functions. Note that there is no known bound on the number of bent rotation symmetric functions.}
  \label{tab:nr_boolean}
  \begin{tabular}{cccccccccccccc}
    &             \\
    \multicolumn{14}{c}{$n$}\\\toprule
    criterion & $4$ & $5$ & $6$ & $7$ & $8$  & $9$ & $10$ & $11$ & $12$ & $13$ & $14$ & $15$ & $16$\\ \midrule
    \# general  & $2^{16}$  & $2^{32}$ & $2^{64}$ & $2^{128}$ & $2^{256}$  & $2^{512}$ & $2^{1024}$ & $2^{2048}$ & $2^{4096}$ & $2^{8192}$ & $2^{16384}$ & $2^{32768}$ & $2^{65536}$\\\midrule
    \# bent & 896 & $-$ & 5425430528 & $-$ & $2^{106.3}$ & $-$ & $2^{638}$ & $-$ & $2^{2510}$ & $-$ & $2^{9908}$ & $-$ & $2^{39203}$\\\midrule
    \# RS & $2^{6}$ & $2^{8}$ & $2^{14}$ & $2^{20}$ & $2^{36}$ & $2^{60}$ & $2^{108}$ & $2^{188}$ & $2^{352}$ & $2^{632}$ & $2^{1182}$ & $2^{2192}$ & $2^{4116}$\\
\bottomrule
  \end{tabular}
\end{table}

Unfortunately, despite belonging to a much smaller class, the space of rotation symmetric Boolean functions still becomes too large for exhaustive search already for $n=9$. This motivates the need to investigate further diverse metaheuristic techniques and the construction of rotation symmetric Boolean functions.

Multiple works leverage evolutionary algorithms to construct Boolean functions with specific properties, commonly focusing on properties like balancedness and nonlinearity, which we also consider in this work. However, most of these studies do not consider rotation symmetric Boolean functions but remain confined to the classes of balanced, highly nonlinear functions or bent functions.
On the other hand, the literature on rotation symmetric Boolean functions and metaheuristics is much more sparse. Despite this dearth of research, some significant findings were made more than 15 years ago~\cite{4167738}. The first work considering evolutionary algorithms in this context appeared only in 2022~\cite{Wang2022}.

This paper investigates how various evolutionary algorithms can construct rotation symmetric Boolean functions, including both bent and balanced functions. We consider three solution encodings: bitstring, tree, and floating point, and two fitness functions. To the best of our knowledge, we are the first to investigate tree and floating-point encodings for this problem. The tree encoding represents an especially intriguing option, as state-of-the-art results indicate its superior performance over bitstring (see, e.g.,~\cite{Djurasevic2023}).
As far as we know, no prior work has applied evolutionary algorithms to construct bent rotation symmetric Boolean functions.
Our main findings are:
\begin{compactitem}
    \item We managed to find rotation symmetric Boolean functions for every tested dimension. Still, we mention that genetic programming (GP) evolving general Boolean functions finds functions with the same or higher nonlinearity. Therefore, we cannot conclude that finding a rotation-symmetric Boolean function is simpler due to the smaller search space. 
    \item While tree encoding is considered the best approach for general Boolean functions, we observe that both bitstring and floating point encodings perform better for rotation symmetric functions. This is because the latter two encodings significantly reduce the search space due to efficient encoding, while this is not the case for GP (tree encoding).
    \item While the best results in related works are reported with customized heuristics, we reached the same (or even better) values with general metaheuristics. As such, we question whether developing custom heuristics is worthwhile compared to, e.g., developing more powerful fitness functions.
\end{compactitem}

\section{Background}
\label{sec:background}

Let us denote positive integers with $n$ and $m$: $n, m \in \mathbb{N}^+$.
Next, we denote the Galois (finite) field with two elements as $\mathbb{F}_{2}$ and the Galois field with $2^n$ elements by $\mathbb{F}_{2^n}$. 
An $(n, m)$-function represents a mapping $F$ from $\mathbb{F}_{2}^{n}$ to $\mathbb{F}_{2}^{m}$. 

When $m=1$, the function $f$ is called a Boolean function (in $n$ inputs/variables). We endow the vector space $\mathbb{F}_2^n$ with the structure of that field, since for every $n$, there exists a field $\mathbb{F}_{2^n}$ of order $2^n$ that is an $n$-dimensional vector space. The usual inner product of $a$ and $b$ equals $a\cdot b = \bigoplus_{i=1}^{n} a_{i}b_{i}$ in $\mathbb F_{2}^n$.

\subsection{Boolean Function Representations}

The simplest way to uniquely represent a Boolean function $f$ on $\mathbb{F}_{2}^{n}$ is by its truth table (TT). The truth table of a Boolean function $f$ is the list of pairs of function inputs (in $ \mathbb F_2^n$) and function values, with the size of the value vector being $2^n$. 
The value vector is the binary vector composed of all $f(x), x \in \mathbb{F}_2^n$, with a certain order selected on $F_2^n$. 
Usually, as seen in, e.g., ~\cite{carlet_2021}, one uses a vector $(f(0),\ldots, f(1))$ that contains the function values of $f$, ordered lexicographically. 
While the truth table representation is simple and ``human-readable'', little can be deduced from it except if the function is balanced (as discussed in Section~\ref{sec:boolean_properties}).

The Walsh-Hadamard transform $W_{f}$ is a unique representation of a Boolean function $f$ that measures the correlation between $f(x)$ and the linear functions $a\cdot x$, see, e.g., ~\cite{carlet_2021}:\footnote{Note that the sum is calculated in ${\mathbb Z}$.}
\begin{equation}
W_{f} (a) = \sum\limits_{x \in \mathbb{F}_{2}^{n}} (-1)^{f(x) + a\cdot x}.
\end{equation}

The Walsh-Hadamard transform is very useful as many Boolean function properties can be evaluated through it. 
Since the complexity of calculating the Walsh-Hadamard transform with a naive approach equals $2^{2n}$, it is common to employ a more efficient method called the fast Walsh-Hadamard transform, where the complexity is reduced to $n2^n$. 

\subsection{Boolean Function Properties and Bounds}
\label{sec:boolean_properties}

\paragraph{Balancedness.}
A Boolean function $f$ is called balanced if it takes the value one exactly the same number of times ($2^{n-1}$) as the value zero when the input ranges over ${\mathbb F}_2^n$.

\paragraph{Nonlinearity.}
The minimum Hamming distance between a Boolean function $f$ and all affine functions, i.e., the functions with the algebraic degree\footnote{The algebraic degree $deg_f$ of a Boolean function $f$ is defined as the number of variables in the largest product term of the function's algebraic normal form having a non-zero coefficient, see, e.g.,~\cite{MacWilliams-Sloane}.} at most 1 (in the same number of variables as $f$), is called the nonlinearity of $f$.
The nonlinearity $nl_{f}$ of a Boolean function $f$ can be easily calculated from the Walsh-Hadamard coefficients, see, e.g.,~\cite{carlet_2021}:
\begin{equation}
\label{eq:nonlinearity}
nl_{f} = 2^{n - 1} - \frac{1}{2}\max_{a \in \mathbb{F}_{2}^{n}} |W_{f}(a)|.
\end{equation}

\noindent
The Parseval relation $\sum\limits_{a\in {\mathbb F}_2^n}W_f(a)^2=2^{2n}$ implies that the arithmetic mean of $W_f(a)^2$ equals $2^n$. 
Since the maximum of $W_f^2 (a)$ is equal to or larger than its arithmetic mean, we can deduce that $\max_{a\in {\mathbb F}_2^n} |W_f(a)|$ must be equal to or larger than $2^{\frac n 2}$. This implies that for every $n$-variable Boolean function, $f$ satisfies the so-called covering radius bound (CRB):
\begin{equation}
\label{eq_boolean_covering}
    nl_{f} \leq 2^{n-1}-2^{\frac n 2 - 1}.
\end{equation}
Eq.~\eqref{eq_boolean_covering} cannot be tight when $n$ is odd. 
For $n$ odd, the bound equals $2\lfloor 2^{n-2}-2^{\frac n 2 - 2}\rfloor$~\cite{568715}.
We will consider Boolean functions that approach the covering radius bound as highly nonlinear. 
We show the values for the covering radius bound for each $n$ in Table~\ref{tab:nl}.

\paragraph{Bent Boolean Functions.}
The functions whose nonlinearity equals the maximal value $2^{n-1}-2^{n/2-1}$ are referred to as bent, and they exist only for $n$ even, see, e.g.,~\cite{carlet_2021}. 
Bent Boolean functions are a very active research topic with applications in, e.g., coding theory~\cite{KERDOCK1972182} and telecommunications~\cite{1056589}.
They are also commonly discussed in cryptography but are not used since they are not balanced (despite being maximally nonlinear). 
\noindent
Bent Boolean functions are rare, and we know the exact numbers of bent Boolean functions for $n\leq 8$ only. 
We also know a naive upper bound $2^{2^{n-1}+\frac{1}{2} {\binom{n}{n/2}}}.$
The numbers of Boolean functions (or upper bound values) are given in Table~\ref{tab:nr_boolean}.

\subsection{Rotation Symmetric Boolean Functions}

A Boolean function over $\mathbb{F}_2^n$ is called rotation symmetric (RS) if invariant under any cyclic shift of input coordinates. Stated differently, it is invariant under a primitive cyclic shift, for instance: 
$$(x_0, x_1, \ldots,  x_{n-1}) \rightarrow (x_{n-1},  x_0, x_1, \ldots, x_{n-2}).$$
Notice that this definition implies that the function $f$ takes the same value for vectors with the same Hamming weight.

Since the above expression holds, the number of rotation symmetric Boolean functions will be less than the number of Boolean functions, as the output value remains the same for certain input values. Let us provide a small example of a rotation symmetric Boolean function when $n=3$.
We obtain the following partitions:
\begin{gather}
\label{eq:partitions}
    \lbrace \left(0,0,0 \right)\rbrace \\\nonumber
    \lbrace \left(0,0,1 \right),\left(0,1,0 \right),\left(1,0,0 \right)\rbrace \\\nonumber
    \lbrace \left( 0,1,1,\right), \left( 1,1,0\right), \left( 1,0,1\right)\rbrace \\\nonumber
    \lbrace \left(1,1,1 \right)\rbrace \\\nonumber
\end{gather}
Thus, four different subsets partition the eight input patterns, and any 3-variable rotation symmetric Boolean function can have a specific value corresponding to each subset. 
An orbit is a rotation symmetric partition composed of vectors equivalent under rotational shifts.

While it is trivial to determine the number of rotation symmetric Boolean functions when $n$ is small, the question remains whether there is a formal way for doing so.
Stanica and Maitra use the Burnside lemma to show that the number of rotation symmetric Boolean functions equals $2^{g_n}$, where $g_n$ equals~\cite{STANICA20081567}:
\begin{equation}
    g_n = \frac{1}{n}\sum_{t|n}\phi(t)2^{\frac{n}{t}},
\end{equation}
where $\phi$ is the Euler phi function.

Bent rotation symmetric functions are maximally nonlinear and invariant under any cyclic shift of input coordinates. Rotation symmetric bent functions are much rarer than general bent
functions~\cite{Mesnager2016}. 
The motivation for considering bent rotation symmetric Boolean functions stems from the fact that such functions can have a simple structure (leading to new bent functions, e.g., Niho bent functions) and representation. Moreover, it is possible to compute them efficiently. However, there are some drawbacks, the most notable being that they are not new, as they belong to the well-known general classes of bent functions~\cite{Mesnager2016}.
We provide results on the upper bounds of nonlinearity and the best-known nonlinearities in Table~\ref{tab:nl}.

\begin{table}
\scriptsize
\setlength{\tabcolsep}{4pt}
  \centering
  \caption{Nonlinearities of Boolean functions. Note that the bound equals the covering radius bound when $n$ is even. Moreover, the best-known nonlinearities when the function is imbalanced and $n$ is even are obtained for bent functions. The best-known results are taken from~\cite{carlet_2021}.}
  \label{tab:nl}
  \begin{tabular}{cccccccccccccc}
    &             \\
    \multicolumn{14}{c}{$n$}\\\toprule
    condition & $4$ & $5$ & $6$ & $7$ & $8$  & $9$ & $10$ & $11$ & $12$ & $13$ & $14$ & $15$ & $16$\\ \midrule
    $2\lfloor 2^{n-2}-2^{\frac n 2 - 2}\rfloor$ & 6 & 12 & 28 & 58 & 120 & 244 & 496 & 1000 & 2016 & 4050 & 8128 & 16292 & 32640\\\midrule
     \multicolumn{14}{c}{balanced}\\\toprule
    best-known $nl_{f}$ & 4 & 12 &  26 & 56 & 116 & 240 & 492 & 992 & 2010 & 4036 & 8120 & 16272 & NA\\\midrule
    \multicolumn{14}{c}{imbalanced}\\\toprule
    best-known $nl_{f}$ & 6 & 12 & 28  & 56 & 120 & 242 & 496 & 996 & 2016 & 4040 & 8128 & 16276 & 32640\\
\bottomrule
  \end{tabular}
\end{table}

More information about Boolean functions and their properties can be found in, e.g.,~\cite{MacWilliams-Sloane,carlet_2021}.

\section{Related Work}
\label{sec:related}

Many works consider metaheuristics and the construction of Boolean functions with specific properties (most often, balancedness and nonlinearity)~\cite{Djurasevic2023}. 
Consequently, we divide this section into two parts. First, we briefly discuss relevant works using evolutionary algorithms to construct bent or balanced and highly nonlinear Boolean functions. Next, we discuss various efforts with metaheuristics to construct rotation symmetric Boolean functions.

The research community has been active in evolving Boolean functions with specific cryptographic properties for almost 30 years~\cite{10.1007/BFb0028471}. While many settings have been tried, the most used solution encodings are the bitstring encoding and the tree encoding~\cite{Djurasevic2023}.
As far as we know, Fuller et al. were the first to consider evolving bent Boolean functions~\cite{FullerDM03}. The authors started with a low-order Boolean function
of input size $n$, and then generated bent functions of higher algebraic order by iteratively adding ANF terms and checking whether the resulting function is bent.
Yang et al. used evolutionary algorithms to evolve bent Boolean functions~\cite{cryptoeprint:2005/322}. They used the trace representation of Boolean functions.
Radek and Vaclav used Cartesian Genetic Programming to evolve bent Boolean functions up to 16 inputs~\cite{10.1007/978-3-319-10762-2_41}. To achieve this goal, the authors used various parallelization techniques. 
Picek and Jakobovic used Genetic Programming to evolve algebraic constructions, which were then used to construct bent Boolean functions~\cite{10.1145/2908812.2908915}. The authors showcased that the approach is highly efficient and provided results for up to 24 inputs, marking the first time that EC successfully constructed such large bent Boolean functions.
Husa and Dobai employed linear genetic programming to evolve bent Boolean functions, reporting superior results compared to related works, as they managed to evolve bent Boolean functions up to $24$ inputs~\cite{10.1145/3067695.3084220}.

Stanica et al. used simulated annealing to evolve rotation symmetric Boolean functions~\cite{StanicaMC04}. By reducing the search space in this manner, the authors could construct $9$-variable plateaued functions with nonlinearity $240$ (among other properties).
Kavut et al. utilized a steepest descent-like iterative algorithm to discover highly nonlinear Boolean functions~\cite{4167738}. The authors found imbalanced Boolean functions in $9$ variables with a nonlinearity of $241$. This represented a significant breakthrough, as the question of whether such functions existed had remained unanswered for nearly three decades. Moreover, the authors found Boolean functions in $10$ variables with nonlinearity $492$.
Kavut and Yucel used a steepest-descent-like iterative algorithm to construct imbalanced Boolean functions in $9$ variables with nonlinearity $242$~\cite{KAVUT2010341}.
Liu and Youssef used simulated annealing to construct balanced rotation symmetric Boolean functions with nonlinearity equal to 488~\cite{4729749}. 
Wang et al. employed genetic algorithms to construct rotation symmetric Boolean functions~\cite{Wang2022}. The authors reported constructing balanced, highly nonlinear rotation symmetric functions.

\section{Experimental Settings}
\label{sec:settings}

\subsection{Representations}
We consider three encodings: bitstring, floating point, and tree-based GP.

\subsubsection{Bitstring Encoding}

The most widely used method for encoding a Boolean function is the bitstring representation~\cite{Djurasevic2023}. The bitstring represents the truth table of the function with which the algorithm works directly.
For a general Boolean function with $n$ inputs, the truth table is encoded as a bit string with a length of $2^n$.
In the case of rotation symmetric Boolean functions, the number of truth table entries that need to be encoded is considerably smaller.
For instance, for a 3-variable function, instead of $2^3 = 8$ bits, we only need to encode 4 bits, which is equal to the number of partitions in the example in the previous section (see Eq.~\eqref{eq:partitions}).
The number of distinct bits that need to be encoded, corresponding to the genotype length, is shown in Table~\ref{tab:genotype} for a given number of variables.

\begin{table}
\setlength{\tabcolsep}{4pt}
  \centering
  \caption{The number of the encoding bits (genotype size) for rotation symmetric Boolean functions}
  \label{tab:genotype}
  \small
  \begin{tabular}{ccccccccccccccccc}
    &             \\
    \toprule
    variables&	1&	2&	3&	4&	5&	6&	7&	8&	9&	10&	11&	12&	13&	14&	15&	16 \\
    $g_n$&	2&	3&	4&	6&	8&	14&	20&	36&	60&	108&	188&	352&	632&	1182&	2192&	4116 \\
\bottomrule
  \end{tabular}
\end{table}

In each evaluation, the bitstring genotype is first decoded into the full Boolean truth table, and the desired property is calculated.
Although the bitstring representation usually performs worse than other encodings~\cite{Djurasevic2023}, especially for a larger number of variables, this might not be the case here due to the largely reduced genotype size.

The corresponding variation operators we use are the simple bit mutation, which inverts a randomly selected bit, and shuffle mutation, which shuffles the bits within a randomly selected substring.
For the crossover operators, we use the one-point crossover, which combines a new solution from the first part of one parent and the second part of the other parent with a randomly selected breakpoint.
The second operator is the uniform crossover that randomly selects one bit from both parents at each position in the child bitstring that is copied.
Each time the evolutionary algorithm invokes a crossover or mutation operation one of the previously described operators is randomly selected.

\subsubsection{Floating Point Encoding}

The second approach we use for representing a Boolean function is the floating point genotype, defined as a vector of continuous variables. 
With this representation, one needs to define the translation of a vector of floating point numbers into the corresponding genotype, which is then translated into a full truth table (binary values). 
The idea behind this translation is that each continuous variable (a real number) of the floating point genotype represents a subsequence of bits in the genotype.
All the real values in the floating point vector are constrained to the interval $[0, 1]$. 
If the genotype size is $g_n$, the number of bits represented by a single continuous variable of the floating point vector can vary and is defined as:
\begin{equation}
\label{eq:decode}
decode = \frac{g_n}{dimension},
\end{equation}
where the parameter $dimension$ denotes the floating point vector size (number of real values). This parameter can be modified as long as the genotype size is divisible by this value. The first step of the translation is to convert each floating point number to an integer value.
Since each real value must represent $decode$ bits, the size of the interval decoding to the same integer value is given as:
\begin{equation}
interval = \frac{1}{decode}.
\end{equation}

To obtain a distinct integer value for a given real number, every element $d_i$ of the floating point vector is divided
by the calculated interval size, generating a sequence of integer values:
\begin{equation}
int\_value_i = \floor*{\frac{d_i}{interval}}.
\end{equation}

The final translation step consists of decoding the integer values to a binary string that can be used for evaluation.
As an example, consider a genotype of $8$ bits. Suppose we want to represent it with $4$ real values; in this case, each real value encodes $2$ bits from the truth table. A string of two bits may have $4$ distinct combinations. Therefore, a single real value must be decoded into an integer value from $0$ to $3$. Since each real value is constrained to $[0, 1]$, the corresponding integer value is obtained by dividing the real value by $2^-2 = 0.25$ and truncating it to the nearest smaller integer. Finally, the integer values are translated into the sequence of bits they encode.

\subsubsection{Tree Encoding}

In the third approach, we use tree-based GP to evolve a function in the symbolic form using a tree representation. 
The terminal set includes a given number of Boolean variables, $x_0$, $x_1$, \ldots, $x_{n-1}$. 
The function set consists of several Boolean primitives that can be used to represent any Boolean function.
In our experiments, we use the following function set: OR, XOR, AND, AND2, XNOR, IF, and function NOT that takes a single argument.
The function AND2 behaves the same as the function AND but with the second input inverted. 
The function IF takes three arguments and returns the second one if the first one evaluates to true and the third one otherwise.
The output of the root node is the output value of the Boolean function. The corresponding truth table of the function $f: \F_2^n \to \F_2$ is determined by evaluating the tree over all possible $2^n$ assignments of the input variables at the leaf nodes. 
The genetic operators used in our experiments with tree-based GP are simple tree crossover, uniform crossover, size fair, one-point, and context preserving crossover~\cite{poli08:fieldguide} (selected at random), and subtree mutation.

Since GP, in this manner, evolves any Boolean function, and not solely rotation symmetric ones, we do not use the GP-derived truth table directly.
Instead, it is treated as the bitstring genotype, the same as in the previous two representations, and decoded into a rotation symmetric function.
This allows GP to use fewer variables than $n$ since the genotype size is considerably smaller than the resulting truth table; for instance, for $n = 8$, the genotype size $g_n = 36$ (instead of $256$), and GP will need to use only $6$ variables to produce a bitstring of at least the required size.
Unfortunately, since the genotype size (see Table~\ref{tab:genotype}) is not a power of 2, a part of the GP-produced bitstring (e.g. of size 64 with six variables) will not be used in any way.
More importantly, there is no direct translation between the truth table of the GP-produced Boolean function, with fewer variables, and the actual rotation symmetric function being decoded and optimized, which may prove detrimental to the GP.

\subsection{Fitness Functions}
\label{subsec:fit}

In our experiments, we optimize two different types of Boolean functions: 1) maximally nonlinear (bent) functions and 2) balanced, highly nonlinear functions.
The first fitness function maximizes the nonlinearity value, $nl_f$, but is designed to consider the whole Walsh-Hadamard spectrum and not only its extreme value (see Eq.~\eqref{eq:nonlinearity}).
More specifically, we count the number of occurrences of the maximal absolute value in the spectrum, denoted as $\#max\_values$.
Since higher nonlinearity corresponds to a \textit{lower} maximal absolute value, we aim for as few occurrences of the maximal value as possible to make it easier for the algorithm to reach the next nonlinearity value.
With this in mind, the fitness function is defined as:
\begin{equation}
\label{eq:bent}
fitness_1 : nl_{f} + \frac{2^n - \#max\_values}{2^n}.
\end{equation}
The second term never reaches the value of $1$ since, in that case, we effectively reach the next nonlinearity level.

With the second criterion, we aim to find balanced, highly nonlinear functions. 
We use a two-stage objective function in which a bonus equal to the previous fitness value is awarded only to a balanced function; otherwise, the objective value is only the balancedness penalty. 
The balancedness penalty $BAL$ is the difference up to the balancedness (i.e., the number of bits to be changed to make the function balanced). 
This difference is included in the objective function with a negative sign to act as a penalty in maximization scenarios.
The delta function $\delta_{BAL, 0}$ assumes the value one when $BAL = 0$ and is zero otherwise. 
\begin{equation}
\label{eq:bal}
fitness_2 : -BAL + \delta_{BAL, 0} \cdot (nl_{f} + \frac{2^n - \#max\_values}{2^n}).
\end{equation}

\section{Experimental Results}
\label{sec:results}

Regarding bitstring (denoted as TT) and tree encoding (denoted as GP), we employ the same evolutionary algorithm: a steady-state selection with a 3-tournament elimination operator. 
In each iteration of the algorithm, three individuals are chosen at random from the population for the tournament, and the worst one in terms of fitness value is eliminated. 
The two remaining individuals in the tournament are used with the crossover operator to generate a new child individual, which then undergoes mutation with individual mutation probability $p_{mut} = 0.5$. Finally, the mutated child takes the place of the eliminated individual in the population.
The population size in all experiments was 500, and the termination criteria were set to $10^6$ evaluations. Each experiment was repeated for $30$ runs.
We consider Boolean function sizes from 8 to 16 inputs, as with less, finding rotation symmetric functions is easy and well within reach of an exhaustive search (see Table~\ref{tab:genotype}).

The floating point representation can be used with any continuous optimization algorithm, which increases its versatility. 
In our experiments, we used the following optimization algorithms: Artificial Bee Colony (ABC)~\cite{karaboga2014comprehensive}, Clonal Selection Algorithm (CLONALG) ~\cite{brownlee2007clonal}, CMA-ES ~\cite{hansen2003reducing}, Differential Evolution (DE) ~\cite{pant2020differential}, Optimization Immune Algorithm (OPTIA)~\cite{cutello2006real}, and a GA-based algorithm with steady-state selection (GA-SST), as described above.

\subsection{General vs Rotation Symmetric Functions}

To facilitate easier comparison with related work, we also provide results for general balanced, highly nonlinear functions and general bent functions, along with the corresponding rotation symmetric ones (Tables~\ref{tab:balanced_best} and~\ref{tab:imbalanced_best}). 
The results for general Boolean functions were reproduced with GP, since in that scenario existing research points to GP as the most efficient approach~\cite{Djurasevic2023,10.1145/3520304.3534087}.
Observe that in the case of balanced functions, the results are better for general functions than for rotation symmetric ones. Our results (the general ones) are also competitive with the best-known nonlinearities up to $n=12$ and for $n=14$ (see Table~\ref{tab:nl}). The nonlinearities when using rotation symmetric functions are the same as the best-known ones only for $n=8, 9$.
The results are slightly different for imbalanced functions (as we do not manage to obtain bent functions in all the cases). For small sizes (up to $n=12$), the results for general functions are better than for rotation symmetric functions, but for $n=14,16$, the opposite is true. We suspect this happens due to the large search space size for such $n$ values, where GP is known to face issues for such large Boolean functions.
The general results are competitive with the best-known nonlinearities up to $n=12$, while the rotation symmetric ones are competitive for $n=8$ only. We note that for general functions, we do not reach bent ones for $n=14,16$; for rotation symmetric ones, bent functions are reached only for $n=8$.

\begin{table}
\scriptsize
\caption{General (30 runs with GP) and rotational symmetric balanced Boolean functions, the best-obtained nonlinearities.}
\label{tab:balanced_best}
\setlength{\tabcolsep}{6pt}
\centering
\begin{tabular}{@{}clllllllll@{}}
\toprule
\multicolumn{10}{c}{Size}                                                                                                                                                                                                           \\ \midrule
   & 8                           & \multicolumn{1}{c}{9} & \multicolumn{1}{c}{10} & \multicolumn{1}{c}{11} & \multicolumn{1}{c}{12} & \multicolumn{1}{c}{13} & \multicolumn{1}{c}{14} & \multicolumn{1}{c}{15} & \multicolumn{1}{c}{16} \\ \midrule
general &   \multicolumn{1}{l}{116} & 240               & 492                 & 992                & 2000                & 4032                & 8120                & 16256                & 32608                  \\ 
rot sym &   \multicolumn{1}{l}{116} & 240               & 488                 & 988                & 1992                & 4012                & 8058                & 16186                & 32456                  \\ \bottomrule
\end{tabular}
\end{table}

\begin{table}
\scriptsize
\caption{General (30 runs with GP) and rotational symmetric imbalanced Boolean functions, the best-obtained nonlinearities.}
\label{tab:imbalanced_best}
\setlength{\tabcolsep}{6pt}
\centering

\begin{tabular}{@{}clllll@{}}
\toprule
\multicolumn{6}{c}{Size}                                                                                                    \\ \midrule
&   8                       & \multicolumn{1}{c}{10} & \multicolumn{1}{c}{12} & \multicolumn{1}{c}{14} & \multicolumn{1}{c}{16} \\ \midrule
general &   \multicolumn{1}{l}{120} & 496                    & 2016                   & 7994                & 32332                  \\ 
rot sym &   \multicolumn{1}{l}{120} & 488                    & 1992                   & 8062                & 32468                  \\ \bottomrule
\end{tabular}
\end{table}

\subsection{Rotation Symmetric Balanced, Highly Nonlinear Boolean Functions}

\begin{figure}
     \centering
     \begin{subfigure}[b]{0.25\textwidth}
         \centering
         \includegraphics[width=\textwidth]{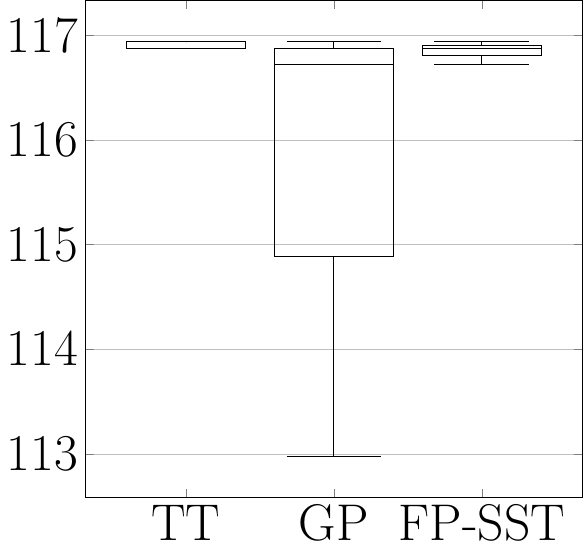}
         \caption{8 variables}
     \end{subfigure}
     \hfill
     \begin{subfigure}[b]{0.25\textwidth}
         \centering
         \includegraphics[width=\textwidth]{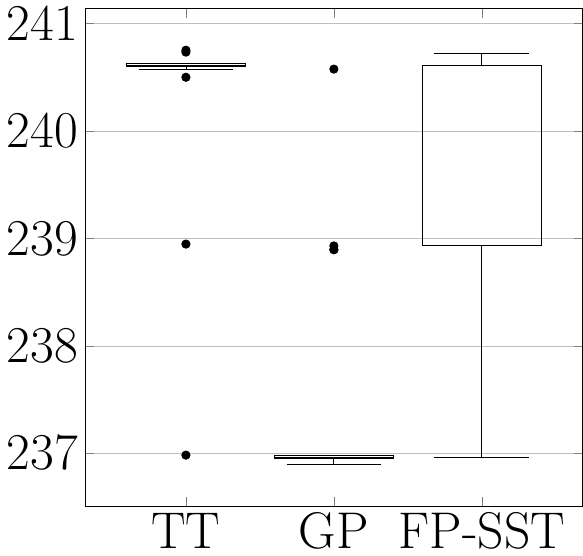}
         \caption{9 variables}
     \end{subfigure}
     \hfill
     \begin{subfigure}[b]{0.25\textwidth}
         \centering
         \includegraphics[width=\textwidth]{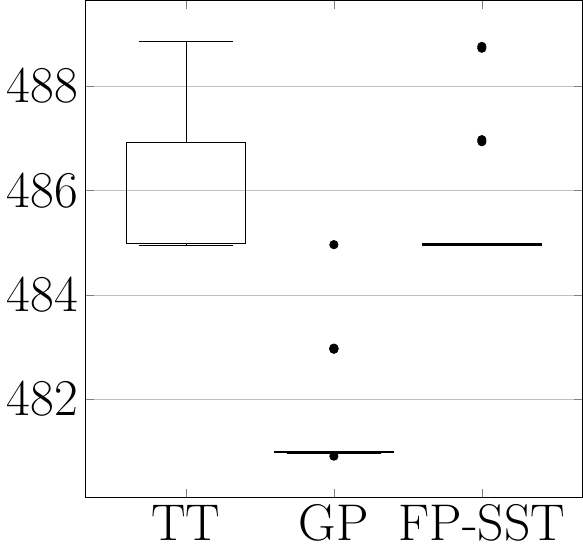}
         \caption{10 variables}
     \end{subfigure}

     \begin{subfigure}[b]{0.24\textwidth}
         \centering
         \includegraphics[width=\textwidth]{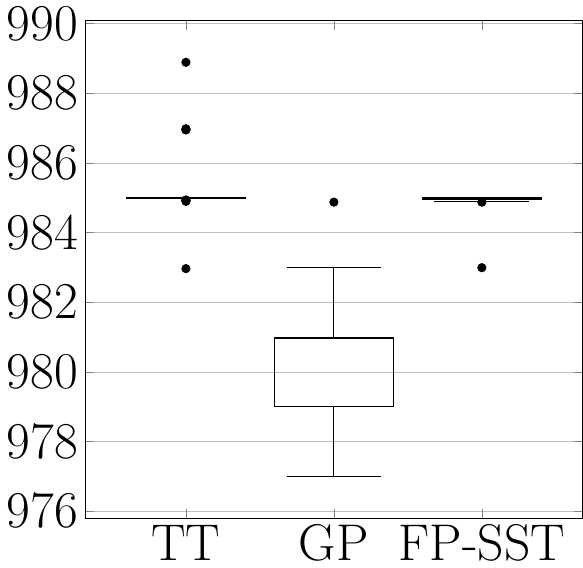}
         \caption{11 variables}
     \end{subfigure}
     \hfill
     \begin{subfigure}[b]{0.25\textwidth}
         \centering
         \includegraphics[width=\textwidth]{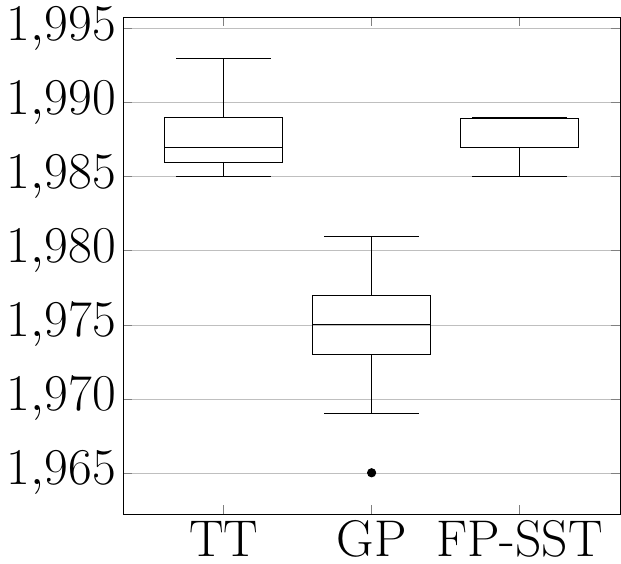}
         \caption{12 variables}
     \end{subfigure}
     \hfill
     \begin{subfigure}[b]{0.25\textwidth}
         \centering
         \includegraphics[width=\textwidth]{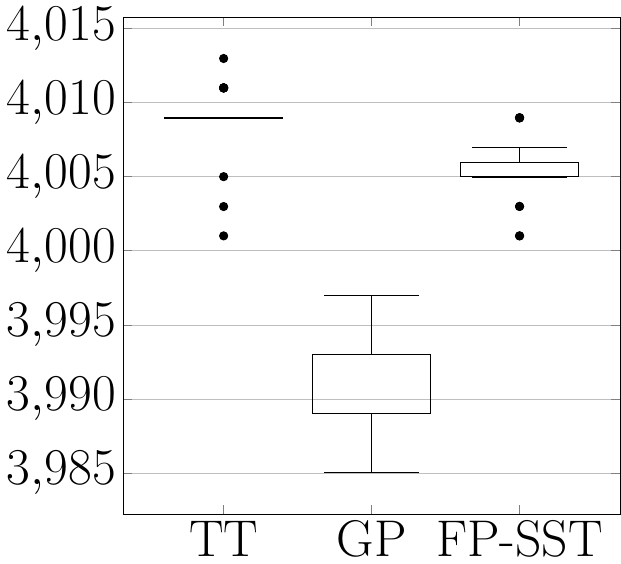}
         \caption{13 variables}
     \end{subfigure}

          \begin{subfigure}[b]{0.25\textwidth}
         \centering
         \includegraphics[width=\textwidth]{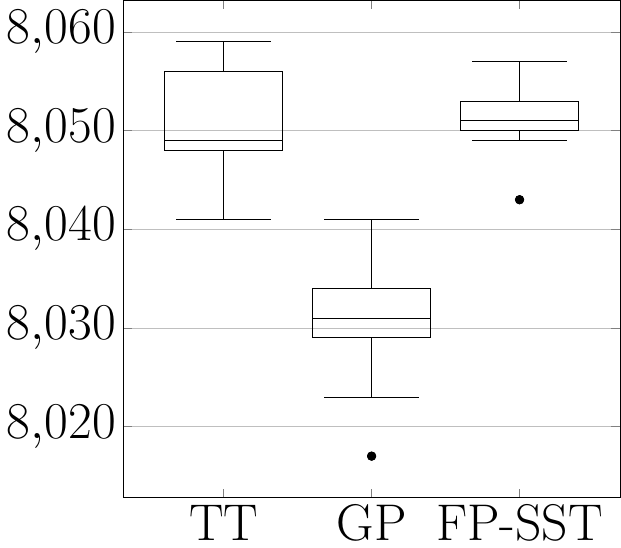}
         \caption{14 variables}
     \end{subfigure}
     \hfill
     \begin{subfigure}[b]{0.25\textwidth}
         \centering
         \includegraphics[width=\textwidth]{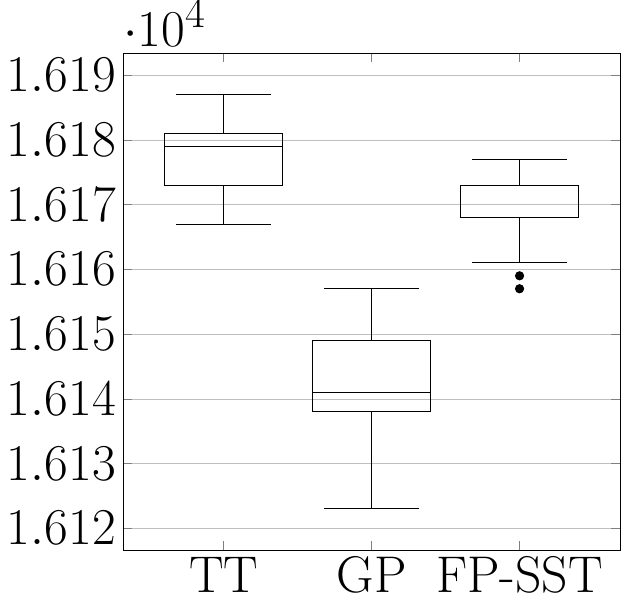}
         \caption{15 variables}
     \end{subfigure}
     \hfill
     \begin{subfigure}[b]{0.25\textwidth}
         \centering
         \includegraphics[width=\textwidth]{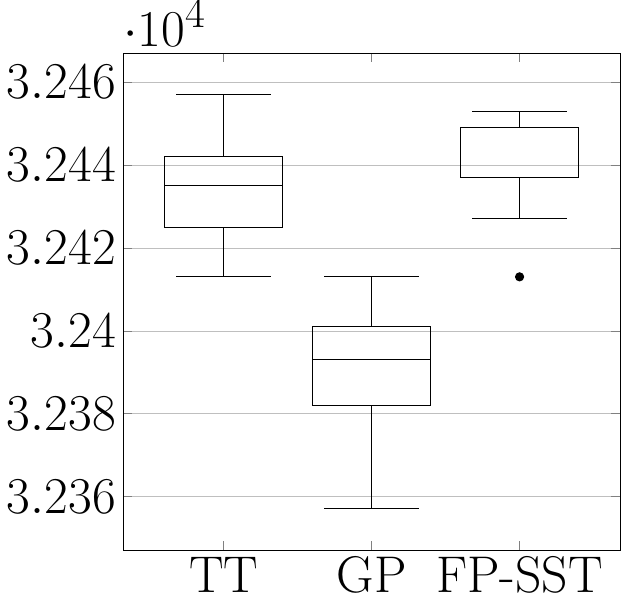}
         \caption{16 variables}
     \end{subfigure}

        \caption{Box plots for nonlinearity values obtained for balanced Boolean functions}
        \label{fig:balanced}
\end{figure}

We provide results for balanced rotation symmetric functions in Table~\ref{tab:median_balanced} and Figure~\ref{fig:balanced}.
Interestingly, the best results for most sizes are attained by the TT representation, except $n=14$ and $n=16$, for which the FP-SST representation provides the best results. 
When FP encoding is used, one can vary the number of bits that a single FP value will represent ($decode$, eq.~\ref{eq:decode}). In our preliminary experiments, the best results were obtained with a relatively small $decode$ (i.e. with one FP value representing a small number of bits), consequently resulting with larger number of FP variables. 
This analysis is not included for brevity, but all FP-based algorithms used the same optimized setting.

\begin{table}
\scriptsize
\caption{Median of nonlinearity values obtained for balanced Boolean functions for different numbers of variables. The N.F. entry denotes that the algorithm could not obtain a balanced Boolean function.}
\label{tab:median_balanced}
\setlength{\tabcolsep}{6pt}
\centering
\begin{tabular}{@{}lccccccccc@{}}
\toprule
\multirow{2}{*}{Representation} & \multicolumn{9}{c}{Size}                                                                                                                                                                                                            \\ \cline{2-10}
                                & 8      & \multicolumn{1}{c}{9}      & \multicolumn{1}{c}{10}     & \multicolumn{1}{c}{11}  & \multicolumn{1}{c}{12}   & \multicolumn{1}{c}{13}   & \multicolumn{1}{c}{14}   & \multicolumn{1}{c}{15}    & \multicolumn{1}{c}{16}    \\ \midrule
TT                              & 116.94 & 240.61 & 484.99 & 985 & 1988 & 4009 & 8049 & 16179 & 32435 \\
GP                              & 116.72 & 236.97 & 480.99 & 981 & 1976 & 3993 & 8032 & 16143 & 32394 \\
FP-ABC                          & 116.69 & 236.95 & 480.99 & 981 & 1977 & 3992 & 8033 & 16147 & 32406 \\
FP-CLONALG                      & 116.88 & 239.73 & 484.98 & 985 & 1988 & 4005 & 8036 & 16137 & 32385 \\
FP-CMAES                        & 116.81 & 236.95 & 480.99 & 977 & 1971 & 3983 & 8014 & 16113 & N.F.    \\
FP-DE                           & 116.80 & 236.93 & 480.98 & 977 & 1969 & 3969 & 7954 & N.F.  & N.F.    \\
FP-OPTIA                        & 115.83 & 237.94 & 484.98 & 985 & 1981 & 3988 & 8019 & 16117 & 32362 \\
FP-SST                          & 116.88 & 240.59 & 484.98 & 985 & 1987 & 4005 & 8053 & 16169 & 32443 \\ \bottomrule
\end{tabular}
\end{table}

\subsection{Rotation Symmetric Bent Boolean Functions}

\begin{figure}
     \centering
     \begin{subfigure}[b]{0.25\textwidth}
         \centering
         \includegraphics[width=\textwidth]{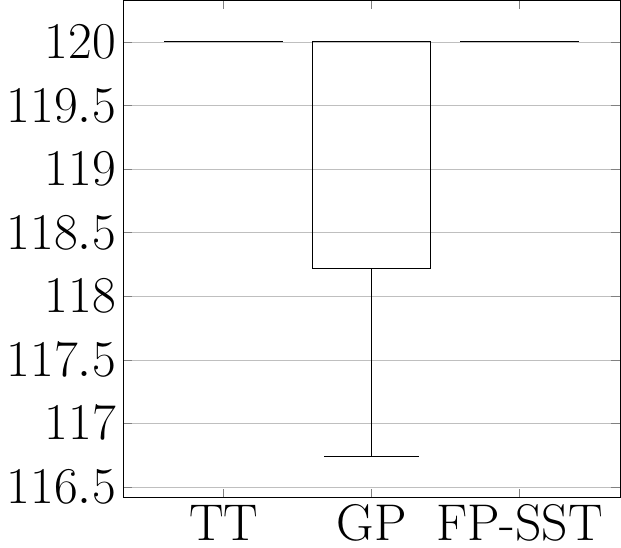}
         \caption{8 variables}
     \end{subfigure}
     \hfill
     \begin{subfigure}[b]{0.237\textwidth}
         \centering
         \includegraphics[width=\textwidth]{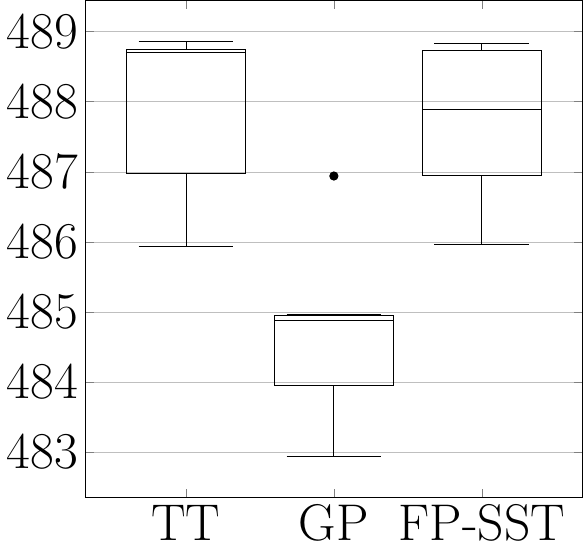}
         \caption{10 variables}
     \end{subfigure}
     \hfill
     \begin{subfigure}[b]{0.25\textwidth}
         \centering
         \includegraphics[width=\textwidth]{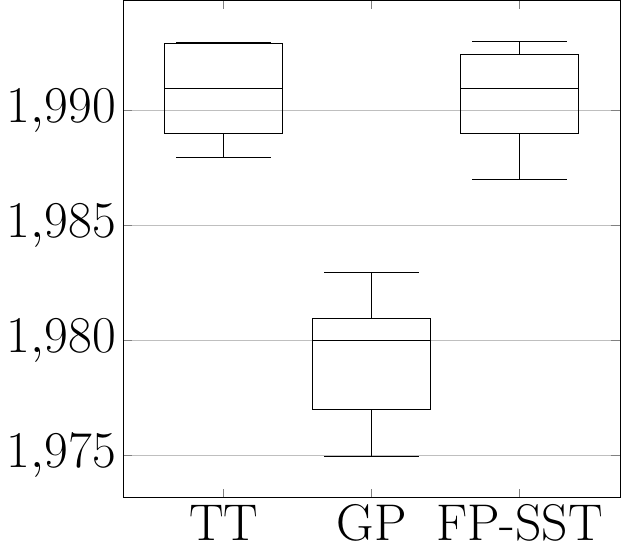}
         \caption{12 variables}
     \end{subfigure}

     \begin{subfigure}[b]{0.25\textwidth}
         \centering
         \includegraphics[width=\textwidth]{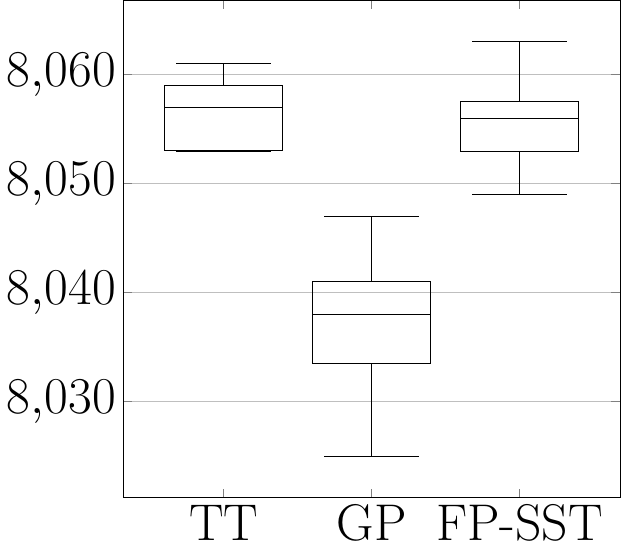}
         \caption{14 variables}
     \end{subfigure}
     \qquad
     \begin{subfigure}[b]{0.24\textwidth}
         \centering
         \includegraphics[width=\textwidth]{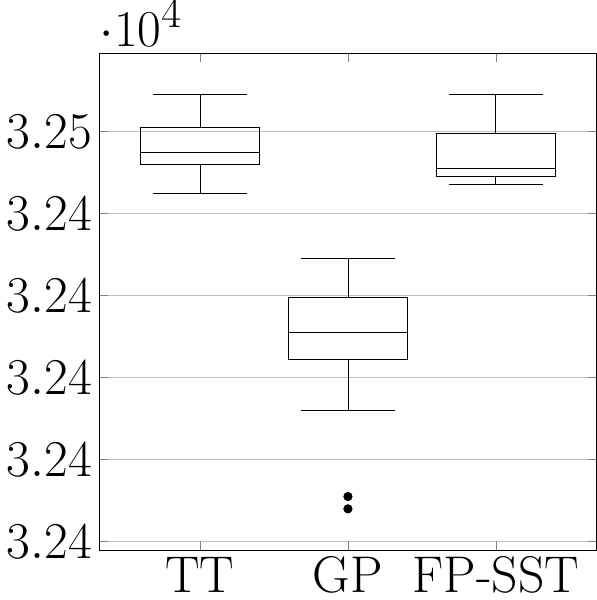}
         \caption{16 variables}
     \end{subfigure}
\hfill
        \caption{Box plots for nonlinearity values obtained for bent Boolean functions}
        \label{fig:bent}
\end{figure}

We provide results for bent (imbalanced) rotation symmetric functions in Table~\ref{tab:median_bent} and Figure~\ref{fig:bent}.
TT provides superior results mainly because of the greatly reduced search space size compared to general Boolean functions. FP-SST is among the best, probably because our implementation includes a variety of floating-point crossover and mutation operators.
Remark that GP provides worse results than TT because there is no semantic link between the GP genotype and the resulting decoded rotation symmetric Boolean function. Among the FP-based algorithms, CMAES and DE exhibit surprisingly unsatisfactory performance, not even managing to find balanced functions for larger $n$ values. We lastly note that the results for rotation symmetric functions are better than general Boolean results for imbalanced nonlinear functions for sizes $14$ and $16$, possibly again because of the reduced search size in the rotation symmetric encoding.

\begin{table}
\scriptsize
\caption{Median of nonlinearity values obtained for bent Boolean functions for a different number of variables.}
\label{tab:median_bent}
\setlength{\tabcolsep}{6pt}
\centering
\begin{tabular}{@{}llllll@{}}
\toprule
\multirow{2}{*}{Representation} & \multicolumn{5}{c}{Size}                                                                                                  \\ \cmidrule(l){2-6} 
                                & \multicolumn{1}{c}{8} & \multicolumn{1}{c}{10} & \multicolumn{1}{c}{12} & \multicolumn{1}{c}{14} & \multicolumn{1}{c}{16} \\ \midrule
TT                              & 120.00                & 488.71                 & 1990.97                & 8056.99                & 32455.50               \\
GP                              & 120.00                & 484.88                 & 1979.99                & 8038.00                & 32411.50               \\
FP-ABC                          & 119.53                & 484.41                 & 1980.00                & 8037.00                & 32410.00               \\
FP-CLONALG                      & 120.00                & 487.89                 & 1990.98                & 8045.00                & 32414.00               \\
FP-CMAES                        & 118.78                & 483.96                 & 1976.00                & 8025.50                & 32382.50               \\
FP-DE                           & 120.00                & 482.98                 & 1974.99                & 8007.50                & 32348.00               \\
FP-OPTIA                        & 119.53                & 486.93                 & 1987.98                & 8036.50                & 32398.50               \\
FP-SST                          & 120.00                & 487.90                 & 1990.96                & 8056.00                & 32458.50               \\ \bottomrule
\end{tabular}
\end{table}

Finally, we compare our results with the two most relevant related works.
Kavut et al. considered rotation symmetric functions in sizes 9 to 11~\cite{4167738}.
For $n=9$, the best nonlinearity for a balanced function equals 240, the same as we achieve.
For $n=10$, Kavut et al. report nonlinearity equal to 488 and 492, but the functions are imbalanced in both cases. We reach balanced functions with nonlinearity 488.
For $n=11$, the authors report a nonlinearity of $988$ for the balanced function and $992$ for the imbalanced function; we also reach the nonlinearity of $988$ for balanced functions.
Later, Kavut et al. applied affine transformation and changed imbalanced functions into balanced ones, but the resulting functions are not rotation symmetric anymore, prohibiting direct comparison. Moreover, to reach such results, they utilize custom heuristics. 

Wang et al. used a custom version of the genetic algorithm (GA) for their experiments and considered only balanced rotation symmetric functions~\cite{Wang2022}. More precisely, they use ``vanilla'' GA, followed by two custom algorithm modifications where good results are reached only for those modified algorithms. For $n=8$, they reach nonlinearity $116$, the same as we. For $n=10$, they obtain a nonlinearity of $488$, which is again the same as we achieve. Finally, for $n=12$, they reported a nonlinearity of $1996$ but only provided an example with nonlinearity $1992$, which is the same as our best result.

\section{Conclusions and Future Work}
\label{sec:conclusions}

This paper explores the difficulty of evolving rotation symmetric Boolean functions. While this class of Boolean functions is much smaller than general Boolean functions, we did not observe the problem to be simpler. Nevertheless, the obtained results are good and rival the related works even though they use customized heuristics while we use generic metaheuristics. Interestingly, we observe that tree encoding is not the best for evolving rotation symmetric functions, but bitstring and floating point work much better (differing from the situation when evolving general Boolean functions). The reason is that the reduction of the search space for bitstring and floating points is significant, while for tree encoding, we can reduce it only marginally.

For future work, we consider two directions to be especially interesting. First, considering (bent) rotation symmetric Boolean functions, it would be interesting to see whether constructions of such functions could be found following the approach from~\cite{10.1145/2908812.2908915}. Indeed, since we observe that all tested techniques struggle with larger Boolean function sizes, circumventing this problem through constructions seems a valid approach. Next, while this work considers rotation symmetric Boolean functions, it would be interesting (and highly relevant from the practical perspective) to consider vectorial rotation symmetric functions (rotation symmetric S-boxes).

\bibliographystyle{abbrv}
\bibliography{bibliography}

\end{document}